# Generating extrema approximation of analytically incomputable functions through usage of parallel computer aided genetic algorithms

Lukasz Swierczewski

luk.swierczewski@gmail.com

Computer Science and Automation Institute, College of Computer Science and Business Administration in Lomza, Poland

This paper presents capabilities of using genetic algorithms to find approximations of function extrema, which cannot be found using analytic ways. To enhance effectiveness of calculations, algorithm has been parallelized using OpenMP library. We gained much increase in speed on platforms using multithreaded processors with shared memory free access. During analysis we used different modifications of genetic operator, using them we obtained varied evolution process of potential solutions. Results allow to choose best methods among many applied in genetic algorithms and observation of acceleration on Yorkfield, Bloomfield, Westmere-EX and most recent Sandy Bridge cores.

## Introduction

Genetic algorithm (GA) is a type of algorithm inspired by the evolution of living organisms in the nature. It belongs to evolution algorithms whose idea was started by John Henry Holland, the American engineer and scientist. GA in a specific way searches in the area of solutions of a problem to find the best solution. The algorithm defines environment in which a specific population of specimens being possible solutions of the problem exists. Next, similarly to organisms in the nature, the specimens are cross-bred, mutated and selection of the best solutions based on the value of adaptation function occurs.

Ideas of genetic algorithm were presented in Fig. 1.

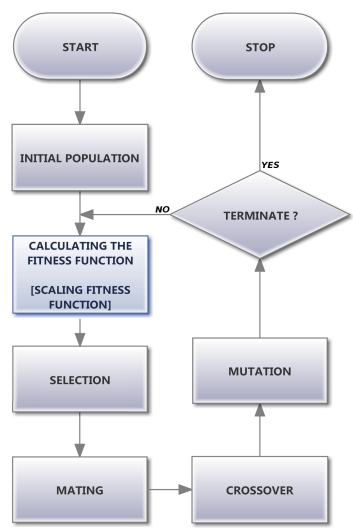

*Fig. 1. Block diagram presenting rule of operation of classic genetic algorithm.*

Genetic algorithms have distinctive features which distinguish them from other methods of calculation:

- genetic operators are used which are adapted to the form of solutions,
- processing of solution population allows to at the same time search in area of solutions from different starting points in order to direct the searching process,
- the quality of current solutions is the sufficient information,
- random elements are introduced on purpose

Genetic algorithms have enormous range of applications in the modern science and engineering. Due to their advantages, they are used when a method of solving problem is not precisely defined, but a method of evaluation of a solution in comparison with others is well known. The classic example is travelling salesman dilemma with whose solution standard algorithms manage relatively poor. GA are also widely used when designing electrical circuits. In such case, evaluation of each specimen is based on the amount of various elements and on their electrical features which can be easily calculated. Using GA, John R. Koza formulated new versions of PID [1] (proportional-integral-derivative controller) which is frequently applied in automatics. Due to the development of FPGA, systems enabling programming structure of electrical circuit contained in them, an experimental project named Golem [2] was designed. It uses genetic algorithms for construction of robots without any support from man. Unfortunately, the project shown that the modern engineering cannot manage such issues and with the use of modern equipment too much time is required for the visible evolution to occur.

Considerably simpler, yet even more popular application of GA is finding extremes of functions. For this purpose also various numerical methods can be used, however, as it turns out, GA often provides good solutions in much shorter amount of time. In some cases, good effects may be obtained by using parallel programming and potential of the modern processors with several arithmetic and logic units. The aspect of genetic algorithms in connection with advanced parallel platforms was brought up in this work.

**Algorithm**

The implementation of GA described and presented in the work is a fragment of Olib library. Its sources written in C/C++ language are available under [3] address. In genetic algorithm the following crossing methods were implemented:

- One-point crossover,
- Two-point crossover,
- Three-point crossover,
- Uniform crossover (Mixing Ratio = 0.5),
- Uniform crossover (Random Mixing Ratio),
- Half Uniform crossover,
- Arithmetic crossover (functions: AND, OR, NOR, NAND, XOR, Random)

One-point crossover method was implemented in two ways. In the first implementation in the event of crossing the result is always one child. In the second implementation with 2/3 probability one child will be born, with 2/9 probability two children and with 1/9 probability three children. Pseudocode representing extension of the method was presented in Listing 1.1.

The most frequently encountered in literature and supported by the algorithm selection methods include:

- Roulette,
- Tournament,
- Linear Ranking

In case of performance testing on various hardware configurations, the condition of stop of algorithm was defined with the number of generations. During the analysis of various genetic methods whose results were presented in Tab. 1, Tab. 2 and Tab. 3 the algorithm ended its operation if the condition was fulfilled:

$$\frac{1}{n}\sum_{i=1}^{n} F(indiv_i) = F(best(indiv))$$

where:
  $F(indiv\_i)$ specifies value of adaptation function for i-specimen,
  $n$ defines the number of specimens in population,
  $F(best(indiv))$ defines value of adaptation function of the best specimen in population.

Generally, it can be said that this condition determines stopping of the program execution when all specimens obtain the same, best value.

---

*Listing 1.1. Pseudocode of the modified version of One-point crossover operator.*

```
void one_point_crossover(individuals_table[], individuals, lvl)
{
    if(lvl < 3)
    {
        random_value = rand();
        if(random_value % 3 == 0)
        {
            lvl++;
            one_point_crossover(individuals_table,
                individuals, lvl);
        }

        // One child is
        // created at this point.
    }
}

one_point_crossover(individuals_table, individuals, 0);
```

---

### Realization calculations

During the calculations the capabilities of parallel programming were used. Genetic operators were made parallel in such a way that they were executed according to the needs parallel on many specimens. The effect of making the executed instructions parallel was obtained through the use of OpenMP library which in a simple way allows to create new threads. This type of software fully utilises capabilities of the modern multi-core processors as well as multiple core platforms with common memory. All calculations were carried out in Linux OS environment and for compilation of programs Intel ICC 12.x version was used.

$$f(x) = \frac{10,2 \cdot x^2 - 0,3 \cdot x + 2}{100} + \sin x + 120 \cdot \sin \frac{x}{10} + 768 \qquad (1)$$

$$f(x) = \begin{cases} \cos x \cdot \sin x \cdot \arctan x \cdot \cos(x+1) + \frac{x}{100} + 5 & \text{if } x < 65 \\ \cos x \cdot \sin x \cdot \arctan x \cdot \cos(x+1) - \frac{x}{100} + 5,65 & \text{if } x \geq 65 \end{cases} \qquad (2)$$

Graphs in the range [2; 130] were presented in Fig. 2 and Fig. 3, respectively. Mainly in this range the focus was put on looking for maximum and minimum of the function. The algorithm was tested on four different calculating platforms:

- Yorkfield: processor Intel Core 2 Quad Q8400
- Bloomfield: processor Intel i7 920
- Westmere-EX: processor Intel Xeon E7-4860
- Sandy Bridge: processor Intel i5-2400

Additionally, during calculations the impact of setting process affinity on the full time of processing data was checked. On Intel i7\ 920 and Xeon E7-4860 processors the Hyper-Threading technology is available which enables increasing performance in result of duplication of some fragments of a processor (mainly registers). The actual impact on the calculations of HT technology was tested. Furthermore, some processors with a code name Sandy Bridge support Turbo Boost 2.0 technology. As it turned out, Intel Core i5-2400 processor using this technology had in some cases problems with achieving full performance.

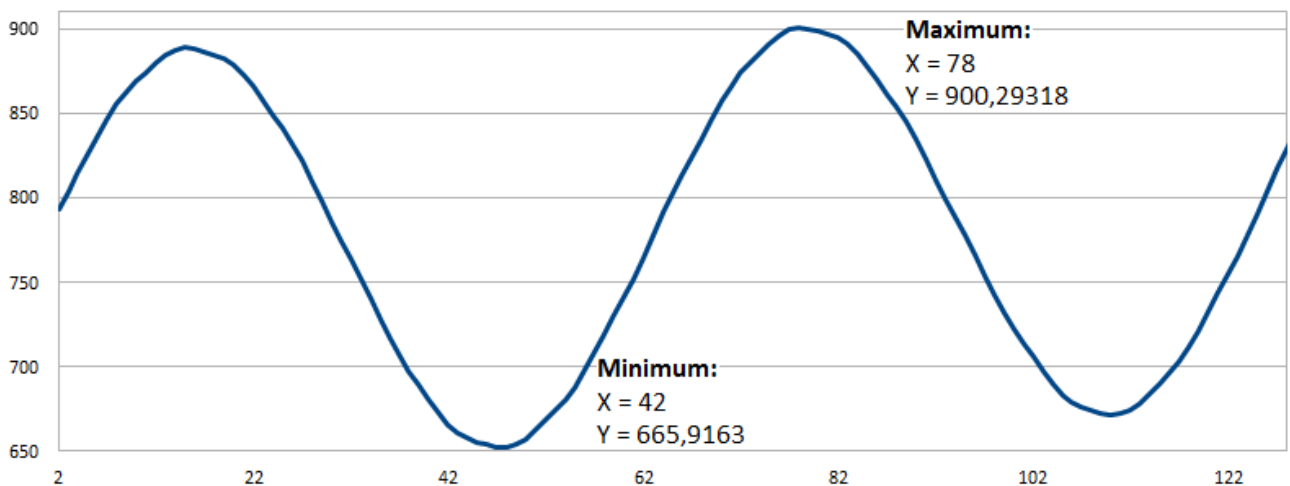

Fig. 2. Values of the function (1) in the interval [2, 130].

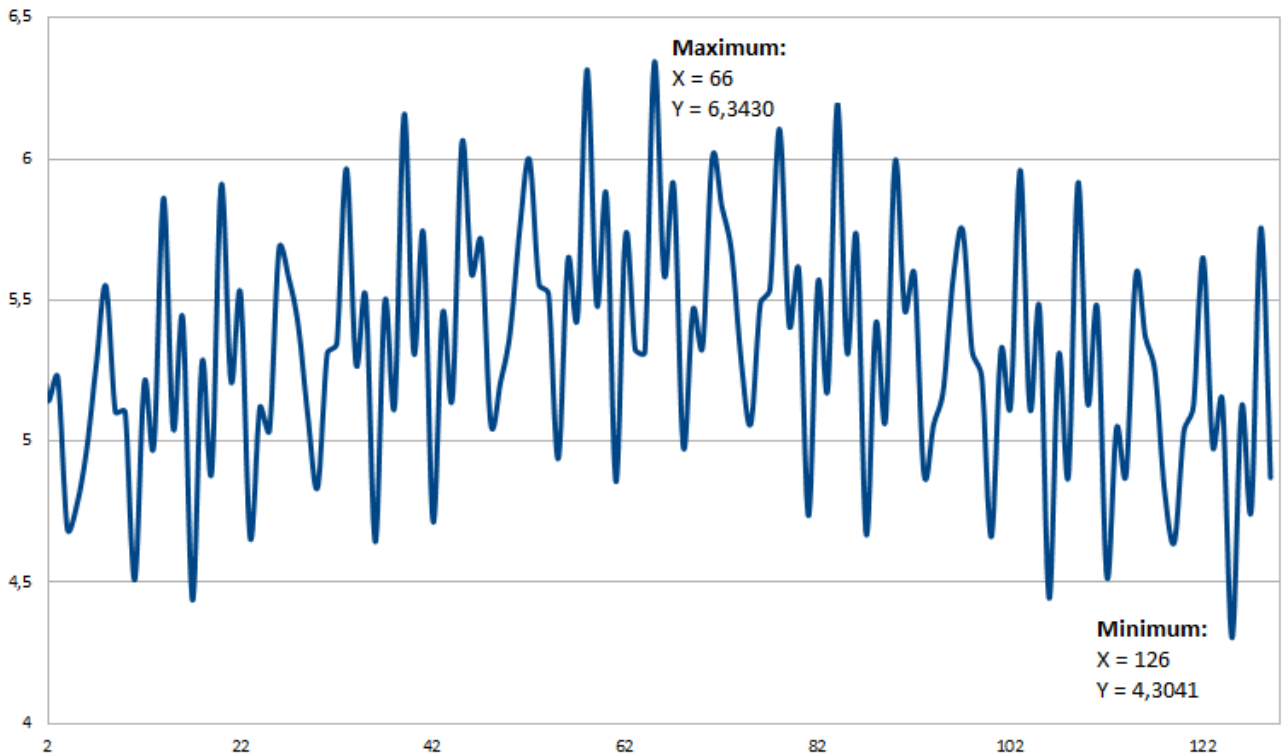

*Fig. 3. Values of the function (2) in the interval [2, 130].*

**The results concerning convergence of algorithm due to the application of various genetic algorithms.**

The produced outcomes regarding capabilities of the implemented genetic algorithms and speed of their convergence for the function (1) were presented in Tab. 1 Analyses for the function (2) were contained in Tab. 2 and Tab. 3.

In case of the first function the genetic algorithm always, regardless of the applied methods, found the appropriate solution. In this instance, convergent outcomes could be very quickly obtained with Half Uniform crossover method, and the algorithm had to generate only 7 generations to fulfil the stop condition. Relatively good results were also obtained as a result of Two-point crossover method. Only 7 generations in case of looking for minimum and 33 in case of maximum is an interesting outcome as well. It is worth to consider results of two different implementations of One-point crossover method. In theory, the version recurrently performed at most three times and generating one to three children should provide better outcomes than the implementation which always creates only one child. However, in case of looking for the maximum it turns out that the situation during the carrying out of the tests looks completely differently and the simpler version of crossing is significantly more quickly convergent. Such result may signify a large impact of random factors on the operation of the entire program.

In Tab. 2 and Tab. 3 considerably more detailed outcomes of the function analysis were presented (2). In Fig. 2 it is possible to observe that the function has many distinctive local maximums and minimums. In this case, genetic algorithm considerably more easily may fall into local minimum or maximum which is not the best result. In conducted tests concerning various selection types the best outcomes were obtained with tournament selection when the size of the group was 10 specimens. In case of smaller tournament groups of 2 specimens, the algorithm was the least frequently finding the correct solution (only 4 times out of 24 trials). Linear Ranking and Roulette methods are solutions marked with indirect effectiveness. Precise results are presented in the respective tables.

In case of function (2) it can also be noticed that different implementations of One-point crossover genetic operator are convergent, in accordance with the predictions, and the version generating according to distribution of probability maximally three children usually causes quicker end of the program operation.

**The results concerning effectiveness of implementation of various genetic operators**

Implementations of various genetic operators are characterised by various calculation complexity. Intuitively, calculation complexity of Two-point crossover operator will be higher than One-point crossover due to division of chromosome in the larger number of points. On the other hand, complexity of all Arithmetic crossover operators will be comparable. Only Arithmetic crossover operating in Random mode (the arithmetic function applied in a case is always random) it may work noticeably slower due to the necessity of formulating a function generating a pseudorandom number.

he time of executing only various functions responsible for crossing in genetic algorithm was presented in Fig 4. A similar comparison for selection and comparison methods was presented in Fig. 5 and Fig. 6, respectively. In Fig. 7 it is possible to see execution of what functions constitutes the operation time of the whole program. According to the measurements, the most time of processor is devoted to selection: about 91%. Other operators of crossing and mutation consume only 3.6% and 1.2%, respectively. It proves that optimisation of genetic operator of selection can yield better gain of total performance.

**The results achieved with the use of various hardware platforms**

Other, significantly more interesting aspect discussed in this work is the use of various hardware platforms for calculations based on genetic algorithms. In case of ordinary processors used at home, an actual acceleration was observed. The outcomes for Intel Core 2 Quad Q8400, i7\ 920 and i5-2400 processors were presented in Fig. 8, Fig. 9 and Fig 10, respectively. The list includes Real Time – the time of program execution from the perspective of user (time of execution of the longest thread) as well as total time calculated for all threads – System Time. On all platforms the tests included execution of algorithm with 1 to 8 threads. Only in respect to Intel i7\ 920 processor an increase of performance in comparison between four and eight threads was observed. This processor has physically only four cores but thanks to Hyper Threading technology it is recognised by the system as a configuration with eight cores. HT technology in case of this processor allows to reduce time by additional 740 seconds, that is about 29.5% in comparison with the ordinary time of use of four cores. The typical fall of performance when comparing times for four and five threads is most probably caused by platform's problems with assigning the excessive, fifth thread, to one of the four physical processors and its migration during operation of the program through various cores. The problem disappears in case of six and higher number of threads where the potential of Hyper-Threading technology is much better utilised. In case of Core 3 Quad Q8400 and i5-2400 processors, the time of execution falls only when four or less threads are used. Further threads created on these platforms are still executed by four processors. They have to be divided into smaller time quanta and properly assigned to processors which causes fall of performance. It is for this reason why usually it is presumed that the number of calculation threads in a program should be equal to the number of available processors in the system. In Fig. 10 an extreme fall of performance in the moment of moving from four to five threads in the case of i5-2400 processor can be seen. When the number of threads does not exceed the physical number of cores the processor works with full clock rate amounting to 3100 MHz. When the number of threads was larger, Turbo Boost 2.0 technology set clock rate to 1600 MHz which caused a considerable fall of performance.

A very advanced platform constructed with four ten-core Intel Xeon 7-4860 processors was used in the tests. These processors support Hyper-Threading technology which makes one unit to be seen by the system as 20 cores. The system is built from four such processors, therefore, in theory it enables running a program with 80 threads. For purely practical reasons performance measurements were carried out using only 1 to 40 threads. The results were presented in Fig. 11. The shortest execution time was obtained when the calculating capabilities of only 19 threads were used. The problem with performance scaling on computers with common memory is frequently that, while the number of processors may increase linearly, speed of operating memory does not increase linearly and it is memory which at some point becomes a bottleneck of the entire configuration.

**The results obtained through changes in process affinity settings**

The process affinity settings were also tested. Process affinity consists of locking processes to specific processors so the process does not have to lose time on migration between processors, during the execution. System planner typically deals with this task. As it turns out, improvement of performance achieved in result of process affinity is slight. For Intel i7 920 processors comparison of times of program execution with enabled and disabled affinity may be seen in Fig. 13. The largest gain was observed when 2 threads were used and it amounted to 106 seconds which is barely 2.9%.

On the other hand, a considerable gain was brought in case of 10 threads and a platform consisting of four Intel Xeon E7-4860 processors. In case of dividing threads to processors by system planner, the time of execution of the program amounted to 1995 seconds. When user manually made all threads to be executed by 10 cores organised within one processor, the time decreased to only 1303. The difference is 34.6% which is a perceptible value. The acceleration was shown in Fig. 12.

*Table. 1. A summary list of speed and correctness of convergence of different crossing methods for function (1).*

| Population size: **64** Mutation type: **Bit inversion** Probability of mutation: **1%** Probability of crossover: **50%** | | | |
|---|---|---|---|
| Crossover type: | | Number of generations: | The best final result: |
| **Function I (Minimum)** | | | |
| Selection type: Linear Ranking | | | |
| One-point crossover : | One child: | 42 | |
| | Max three children: | 17 | |
| Two-point crossover: | | 7 | |
| Uniform crossover: | Mixing Ratio = 0.5: | 7 | |
| | Random Mixing Ratio: | 33 | |
| Half Uniform crossover: | | 7 | **Yes** |
| Arithmetic crossover: | AND: | 61 | |
| | OR: | 15 | |
| | NOR: | 6 | |
| | NAND: | 19 | |
| | XOR: | 13 | |
| | Random: | 6 | |
| **Function I (Maximum)** | | | |
| Selection type: Linear Ranking | | | |
| One-point crossover: | One child: | 21 | |
| | Max three children: | 82 | |
| Two-point crossover: | | 33 | |
| Uniform crossover: | Mixing Ratio = 0.5: | 70 | |
| | Random Mixing Ratio: | 71 | |
| Half Uniform crossover: | | 7 | **Yes** |
| Arithmetic crossover: | AND: | 8 | |
| | OR: | 61 | |
| | NOR: | 63 | |
| | NAND: | 34 | |
| | XOR: | 15 | |
| | Random: | 74 | |

Table 2. *A summary list of speed and correctness of convergence of different crossing and selection methods for function (2).*

| Crossover type: | | Number of generations: | The best final result: |
|---|---|---|---|
| Population size: **64** Mutation type: **Bit inversion** Probability of mutation: **1%** Probability of crossover: **50%** | | | |
| **Function II (Minimum)** | | | |
| Selection type: Linear Ranking / Roulette | | | |
| One-point crossover: | One child: | 14 / 25 | **Yes** / No |
| | Max three children: | 14 / 17 | No / No |
| Two-point crossover: | | 9 / 56 | No / **Yes** |
| Uniform crossover: | Mixing Ratio = 0.5: | 6 / 49 | No / **Yes** |
| | Random Mixing Ratio: | 32 / 15 | No / No |
| Half Uniform crossover: | | 64 / 31 | **Yes** / No |
| Arithmetic crossover: | AND: | 7 / 13 | No / **Yes** |
| | OR: | 6 / 116 | No / **Yes** |
| | NOR: | 23 / 16 | **Yes** / No |
| | NAND: | 6 / 23 | No / No |
| | XOR: | 63 / 81 | **Yes** / No |
| | Random: | 29 / 86 | **Yes** / No |
| **Function II (Maximum)** | | | |
| Selection type: Linear Ranking / Roulette | | | |
| One-point crossover: | One child: | 7 / 105 | No / **Yes** |
| | Max three children: | 6 / 94 | No / No |
| Two-point crossover: | | 65 / 12 | No / **Yes** |
| Uniform crossover: | Mixing Ratio = 0.5: | 15 / 6 | No / No |
| | Random Mixing Ratio: | 7 / 5 | No / No |
| Half Uniform crossover: | | 58 / 7 | **Yes** / No |
| Arithmetic crossover: | AND: | 49 / 9 | No / No |
| | OR: | 12 / 6 | No / **Yes** |
| | NOR: | 6 / 29 | No / No |
| | NAND: | 32 / 148 | No / No |
| | XOR: | 12 / 87 | **Yes** / No |
| | Random: | 7 / 71 | **Yes** / **Yes** |

*Table 3. A summary list of speed and correctness of convergence of different crossing and selection methods for function (2).*

| Population size: **64** Mutation type: **Bit inversion** Probability of mutation: **1%** Probability of crossover: **50%** | | | |
|---|---|---|---|
| Crossover type: | | Number of generations: | The best final result: |
| **Function II (Minimum)** | | | |
| Selection type: Tournament (tournament group: 10) / (tournament group: 2) | | | |
| One-point crossover: | One child: | 5 / 10 | **Yes** / No |
| | Max three children: | 5 / 10 | No / No |
| Two-point crossover: | | 3 / 7 | No / No |
| Uniform crossover: | Mixing Ratio = 0.5: | 3 / 9 | **Yes** / **Yes** |
| | Random Mixing Ratio: | 3 / 8 | No / No |
| Half Uniform crossover: | | 3 / 32 | **Yes** / No |
| Arithmetic crossover: | AND: | 3 / 7 | No / No |
| | OR: | 3 / 9 | No / No |
| | NOR: | 3 / 9 | **Yes** / No |
| | NAND: | 3 / 8 | **Yes** / No |
| | XOR: | 4 / 9 | No / **Yes** |
| | Random: | 3 / 10 | No / **Yes** |
| **Function II (Maximum)** | | | |
| Selection type: Tournament (tournament group: 10) / (tournament group: 2) | | | |
| One-point crossover: | One child: | 3 / 8 | No / No |
| | Max three children: | 3 / 9 | **Yes** / No |
| Two-point crossover: | | 3 / 9 | **Yes** / **Yes** |
| Uniform crossover: | Mixing Ratio = 0.5: | 3 / 13 | **Yes** / No |
| | Random Mixing Ratio: | 3 / 7 | **Yes** / No |
| Half Uniform crossover: | | 3 / 10 | No / No |
| Arithmetic crossover: | AND: | 3 / 9 | No / No |
| | OR: | 3 / 9 | **Yes** / No |
| | NOR: | 3 / 9 | **Yes** / No |
| | NAND: | 3 / 10 | No / No |
| | XOR: | 3 / 15 | No / No |
| | Random: | 3 / 10 | No / No |

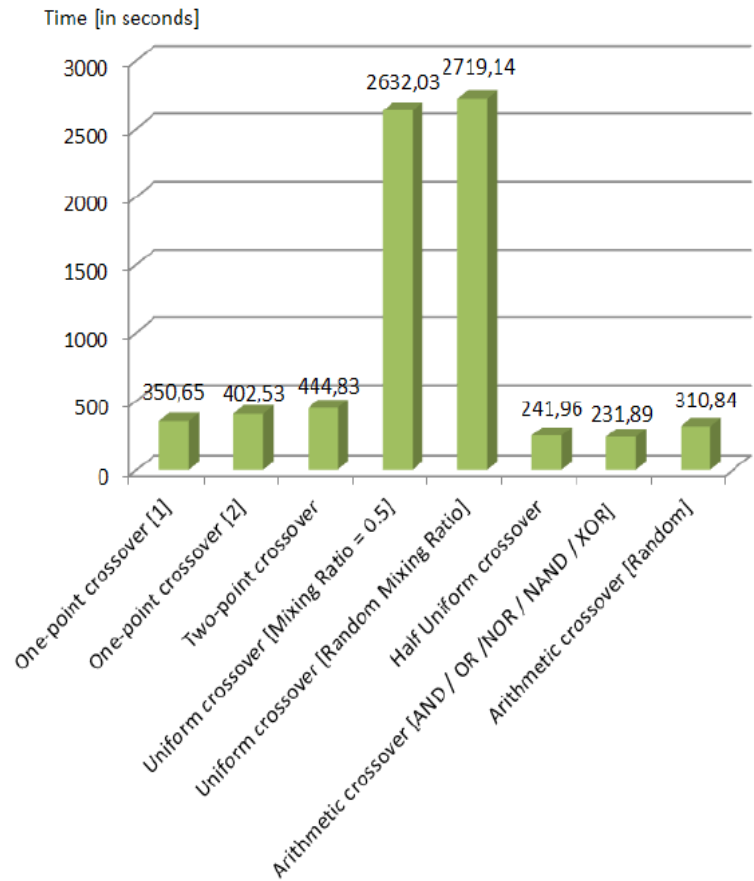

*Fig. 4. Summary comparison of the time of execution of various crossing operators for the algorithm run with the parameters:*
Parameters of the program: Function (2); Range: [2; 1048578]; Searching Maximum; Population size: 16384; Mutation: Bit inversion (probability = 1%); Probability of crossover = 50%; Selection: Roulette; Linear scaling; 10000 generations.

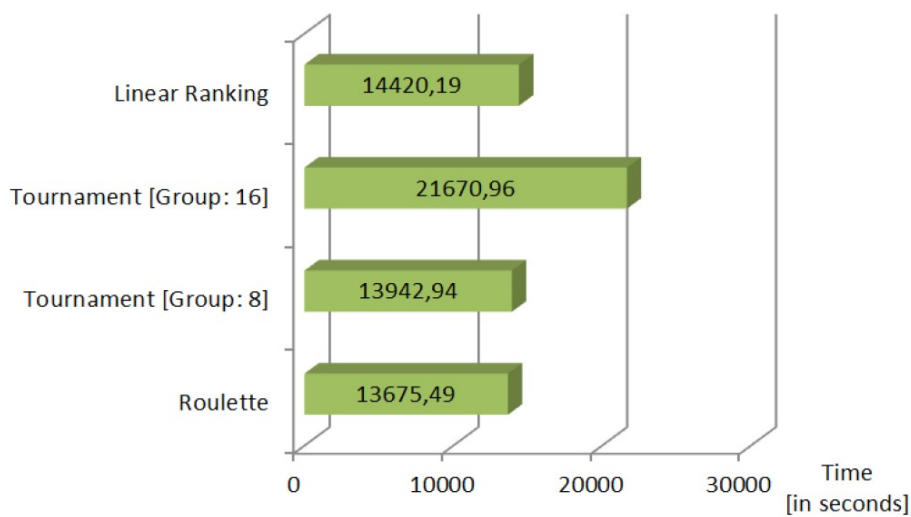

*Fig. 5. Summary comparison of the time of execution of various selection operators for the algorithm run with the parameters:*
Parameters of the program: Function (2); Range: [2; 1048578]; Searching Maximum; Population size: 16384; Mutation: Bit inversion (probability = 1%); Crossover: Two-point crossover (probability = 50%); Linear scaling; 10000 generations.

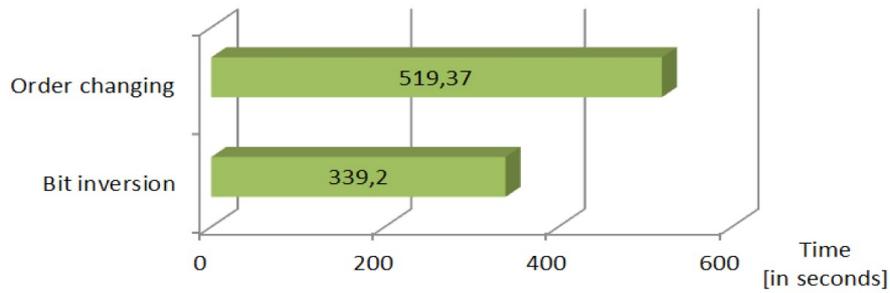

Fig. 6. Summary comparison of the time of execution of various mutation operators for the algorithm run with the parameters:
*Parameters of the program: Function (2); Range: [2; 1048578]; Searching Maximum; Population size: 16384; Probability of mutation = 100%; Crossover: Two-point crossover (probability = 50%); Selection: Roulette; Linear scaling; 10000 generations.*

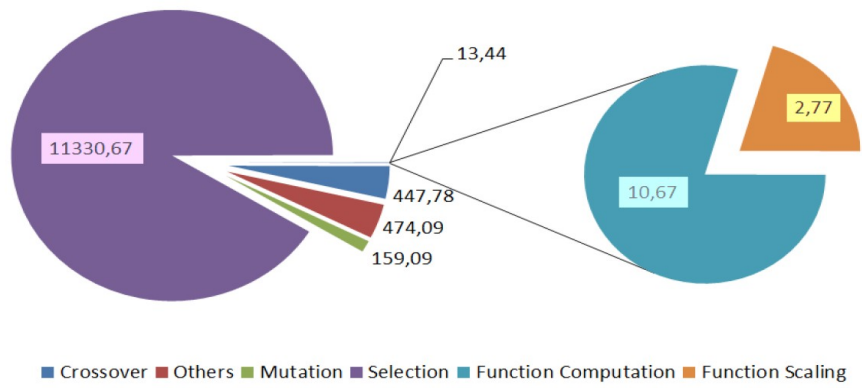

Fig. 7. Execution time for various functions in the program.
*Parameters of the program: Function (2); Range: [2; 1048578]; Searching Maximum; Population size: 16384; Mutation: Bit inversion (probability = 1%); Crossover: Two-point crossover (probability = 50%); Selection: Roulette; Linear scaling; 10000 generations.*
*System time measured for eight threads processor Intel Core i7 920.*

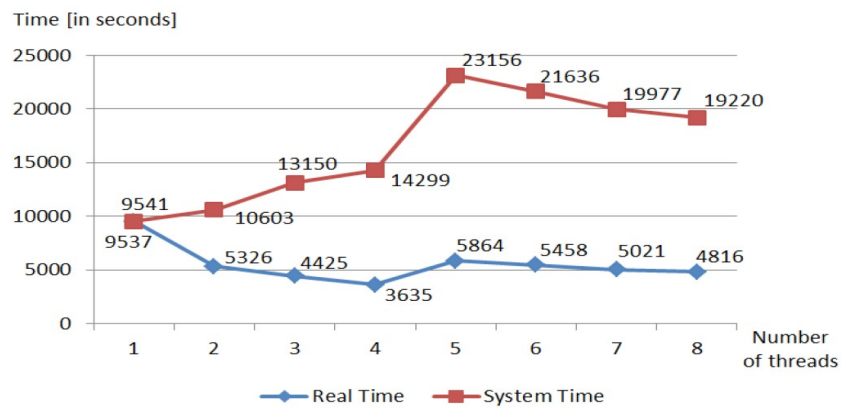

Fig. 8. Acceleration obtained with the use of parallel programming and capabilities of Intel Core 2 Quad Q8400 processor "Yorkfield".
*Parameters of the program: Function (2); Range: [2; 1048578]; Searching Maximum; Population size: 16384; Mutation: Bit inversion (probability = 1%); Crossover: Two-point crossover (probability = 50%); Selection: Roulette; Linear scaling; 10000 generations.*

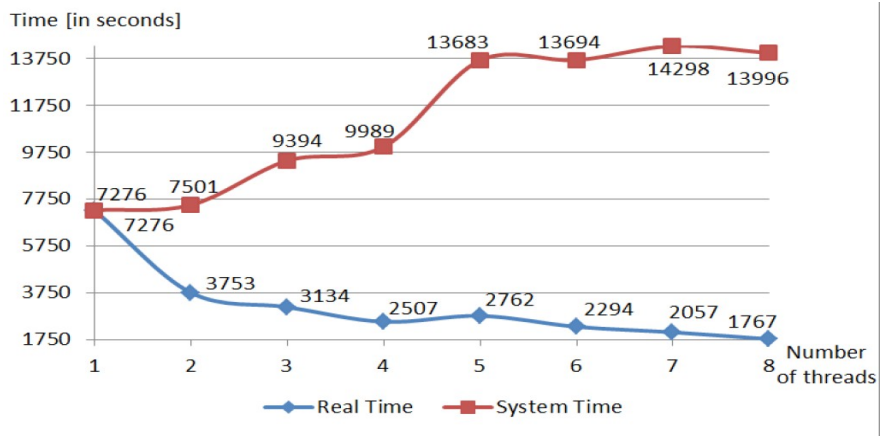

*Fig. 9. Acceleration obtained with the use of parallel programming and capabilities of Intel Core i7 920 processor "Bloomfield".*
*Parameters of the program: Function (2); Range: [2; 1048578]; Searching Maximum; Population size: 16384; Mutation: Bit inversion (probability = 1%); Crossover: Two-point crossover (probability = 50%); Selection: Roulette; Linear scaling; 10000 generations.*

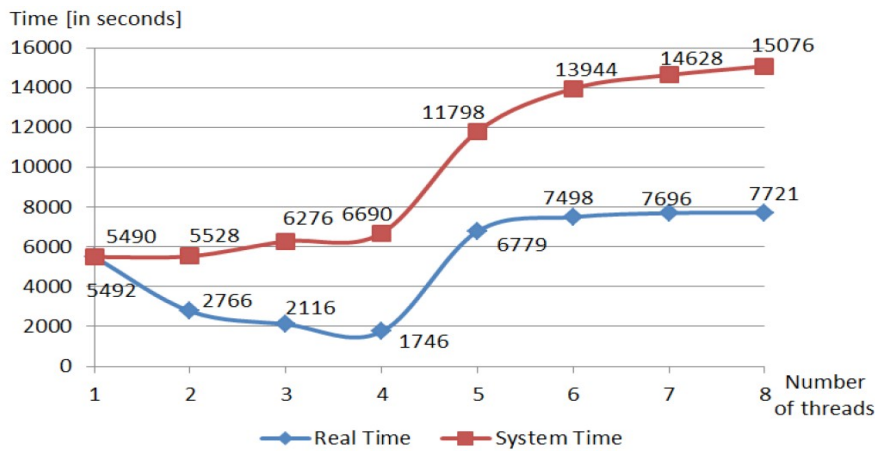

*Fig. 10. Acceleration obtained with the use of parallel programming and capabilities of Intel Core i5-2400 processor "Sandy Bridge".*
*Parameters of the program: Function (2); Range: [2; 1048578]; Searching Maximum; Population size: 16384; Mutation: Bit inversion (probability = 1%); Crossover: Two-point crossover (probability = 50%); Selection: Roulette; Linear scaling; 10000 generations.*

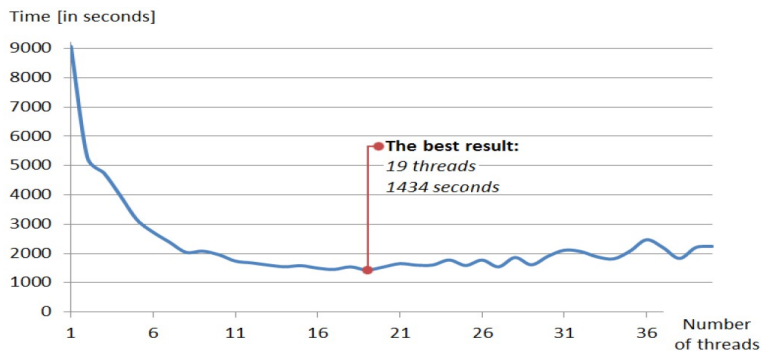

*Fig. 11. Acceleration obtained with the use of parallel programming and capabilities of Intel Xeon E7-4860 processor "Westmere-EX".*
*Parameters of the program: Function (2); Range: [2; 1048578]; Searching Maximum; Population size: 16384; Mutation: Bit inversion (probability = 1%); Crossover: Two-point crossover (probability = 50%); Selection: Roulette; Linear scaling; 10000 generations.*

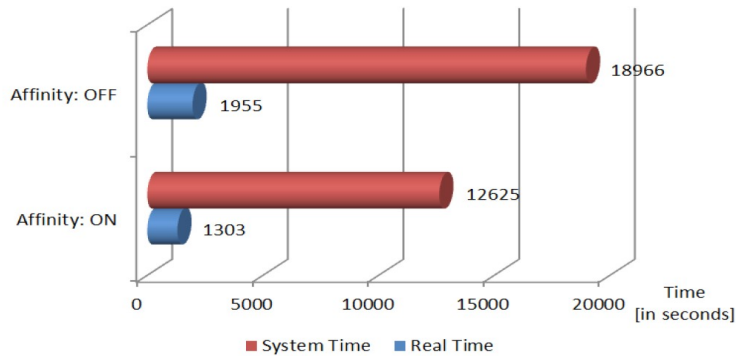

*Fig. 12. Acceleration obtained in result of the application of process affinity on Intel Xeon E7-4860 processor "Westmere-EX".*
*Parameters of the program: Function (2); Range: [2; 1048578]; Searching Maximum; Population size: 16384; Mutation: Bit inversion (probability = 1%); Crossover: Two-point crossover (probability = 50%); Selection: Roulette; Linear scaling; 10000 generations.*

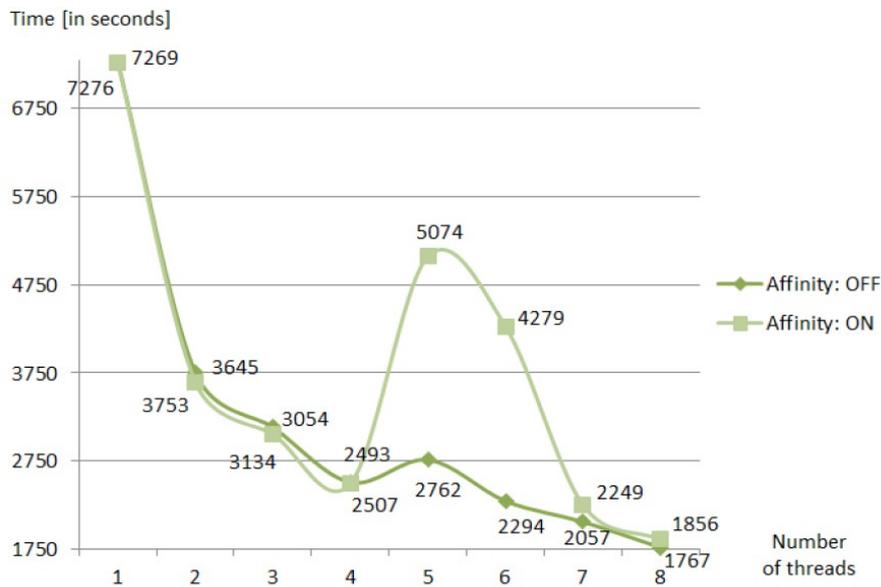

*Fig. 13. Acceleration obtained in result of the application of process affinity on Intel Core i7 920 processor "Bloomfield".*
*Parameters of the program: Function (2); Range: [2; 1048578]; Searching Maximum; Population size: 16384; Mutation: Bit inversion (probability = 1%); Crossover: Two-point crossover (probability = 50%); Selection: Roulette; Linear scaling; 10000 generations.*

## Conclusion

Genetic operators analysed in the article can be divided into more and less demanding methods in respect to the executed operations. As regards crossing, Uniform crossover implementation proved to be the most complex, and in relation to selection this tournament method whose complexity raises with the increase of size of tournament group.

Genetic algorithms relatively well undergo process of paralleling. The acceleration is perfectly noticeable on the modern multi-core processors. In case of more advanced platforms built from many multi-core processors problems with appropriate performance scaling may occur.

The application of programmable graphic accelerators which in the recent years have become very popular may turn out to be a very interesting prospect. Properly utilised, they will certainly allow to even more quickly process data.

## Acknowledgment

The work has been prepared using the supercomputer resources provided by the Faculty of Mathematics, Physics and Computer Science of the Maria Curie-Skłodowska University in Lublin and Computer Science and Automation Institute of the College of Computer Science and Business Administration in Lomza.

# Generacja przybliżeń ekstremów funkcji nieobliczalnych analitycznie dzięki zastosowaniu algorytmów genetycznych ze wsparciem komputerów równoległych

Praca prezentuje możliwości zastosowania algorytmów genetycznych do odnajdywania przybliżeń ekstremów funkcji, których nie można obliczyć w sposób analityczny. Aby zwiększyć efektywność prowadzonych obliczeń algorytm poddano równoleglizacji z wykorzystaniem biblioteki OpenMP. Uzyskano dzięki temu zauważalne przyśpieszenie na platformach o swobodnym dostępie do pamięci wspólnej wykorzystujących procesory wielordzeniowe. Podczas analiz wykorzystano różne modyfikacje operatorów genetycznych, dzięki którym uzyskano zróżnicowane procesy ewolucji osobników, będących potencjalnymi rozwiązaniami. Wyniki umożliwiają wybór najlepszych metod spośród wielu stosowanych w algorytmach genetycznych oraz obserwację akceleracji na układach Yorkfield, Bloomfield, Westmere-EX oraz najnowocześniejszych Sandy Bridge.

Słowa kluczowe: sztuczna inteligencja, algorytmy genetyczne, algorytmy równoległe, ekstrema funkcji